\pdfoutput=1

\documentclass[11pt]{article}

\usepackage[final]{acl}

\usepackage{times}
\usepackage{latexsym}

\usepackage[T1]{fontenc}

\usepackage[utf8]{inputenc}
\usepackage{todonotes}

\usepackage{microtype}

\usepackage{inconsolata}

\usepackage{graphicx}

\usepackage[utf8]{inputenc} 
\usepackage[T1]{fontenc}    
\usepackage{url}            
\usepackage{booktabs}       
\usepackage{amsfonts}       
\usepackage{nicefrac}       
\usepackage{microtype}      
\usepackage{bm}

\usepackage{times}
\usepackage{latexsym}
\usepackage{helvet}  
\usepackage{courier}  
\usepackage{graphicx}
\usepackage{subfigure}  

\usepackage{color}
\usepackage{multirow}
\usepackage{bbm}
\usepackage{algorithm,algorithmic}
\usepackage{float}
\usepackage{caption}
\usepackage{wrapfig}
\usepackage{amsmath}
\usepackage{lipsum}
\usepackage{makecell}
\usepackage{listings}

\newcommand{\Mmat}{{\bf M}}

\newcommand{\Fcal}{\mathcal{F}}

\newcommand{\Rcal}{\mathcal{R}}

\usepackage{tcolorbox}
\tcbuselibrary{listings,skins,breakable}

\newtcblisting{PromptBox}{
  enhanced,
  breakable,                 
  colback=gray!10,           
  boxrule=0pt,
  frame hidden,
  arc=4pt,                   
  left=6pt, right=6pt, top=6pt, bottom=6pt,
  listing only,
  listing options={
    basicstyle=\scriptsize\ttfamily,  
    columns=fullflexible,             
    breaklines=true,                  
    breakatwhitespace=false,          
    keepspaces=true,                  
    showstringspaces=false,
  }
}
\newcommand{\system}{%
 {\sc Pensieve}
}

\title{Memory-QA: Answering Recall Questions Based on Multimodal Memories}

\author{Hongda Jiang\thanks{These authors contributed equally to this work.}\textsuperscript{$\dagger$}, Xinyuan Zhang\footnotemark[1]\textsuperscript{$\dagger$}, Siddhant Garg, Rishab Arora,\\
{\bf Shiun-Zu Kuo, Jiayang Xu, Ankur Bansal, Christopher Brossman, Yue Liu,} \\
{\bf Aaron Colak, Ahmed Aly, Anuj Kumar, \and Xin Luna Dong\textsuperscript{$\dagger$}} \\
Meta Reality Labs \\
\textsuperscript{$\dagger$}\{jhd,dylanz426,lunadong\}@meta.com\\}

\begin{document}
\maketitle

\begin{abstract}

We introduce Memory-QA, a novel real-world task that involves answering recall questions about visual content from previously stored multimodal memories. This task poses unique challenges, including the creation of task-oriented memories, the effective utilization of temporal and location information within memories, and the ability to draw upon multiple memories to answer a recall question. To address these challenges, we propose a comprehensive pipeline, \system, integrating memory-specific augmentation, time- and location-aware multi-signal retrieval, and multi-memory QA fine-tuning. We created a multimodal benchmark to illustrate various real challenges in this task, and show the superior performance of \system over state-of-the-art solutions (up to 14\% on QA accuracy). 
\end{abstract}

\section{Introduction}
\label{sec:intro}

Envision a smart personal assistant capable of persistently remembering events from an individual's life---under explicit user permission---and answer recall questions like {\em “Where did I park my car?” “I had some very good Korean hotpot a while back but which restaurant was that?” “Does this skirt have lower price than the similar one I saw at Macy’s yesterday?”}
This vision dates back to Vannevar Bush’s seminal concept of {\em MEMEX (MEMory \& EXpansion)}~\citep{bush1945we}, and has recently been revitalized under the emerging paradigm of the {\em Second Brain}~\citep{forte2022building}. 

Achieving this vision introduces a number of technical challenges, including memory recording, compression, storage, and search. In this paper, we take an initial step towards this long-term vision by addressing a simpler yet essential sub-problem: {\em recording memory snapshots on user request and enabling question answering over these recorded memories}. For instance, a user may issue invocation commands such as {\em "remember my parking lot," "remember this restaurant", "remember this dress"}, or more generally, {\em "remember this"} through wearable devices, mobile phones, etc. In response, the assistant captures the user's intent along with a visual snapshot of their current view. These on-demand, user-initiated memory recordings serve as the foundation for answering future memory-related questions. We refer to this task as {\em Memory-QA}, comprised of multimodal memory recording, memory retrieval, and question answering.

\begin{figure}[t!]
\centering
\includegraphics[width=\linewidth]{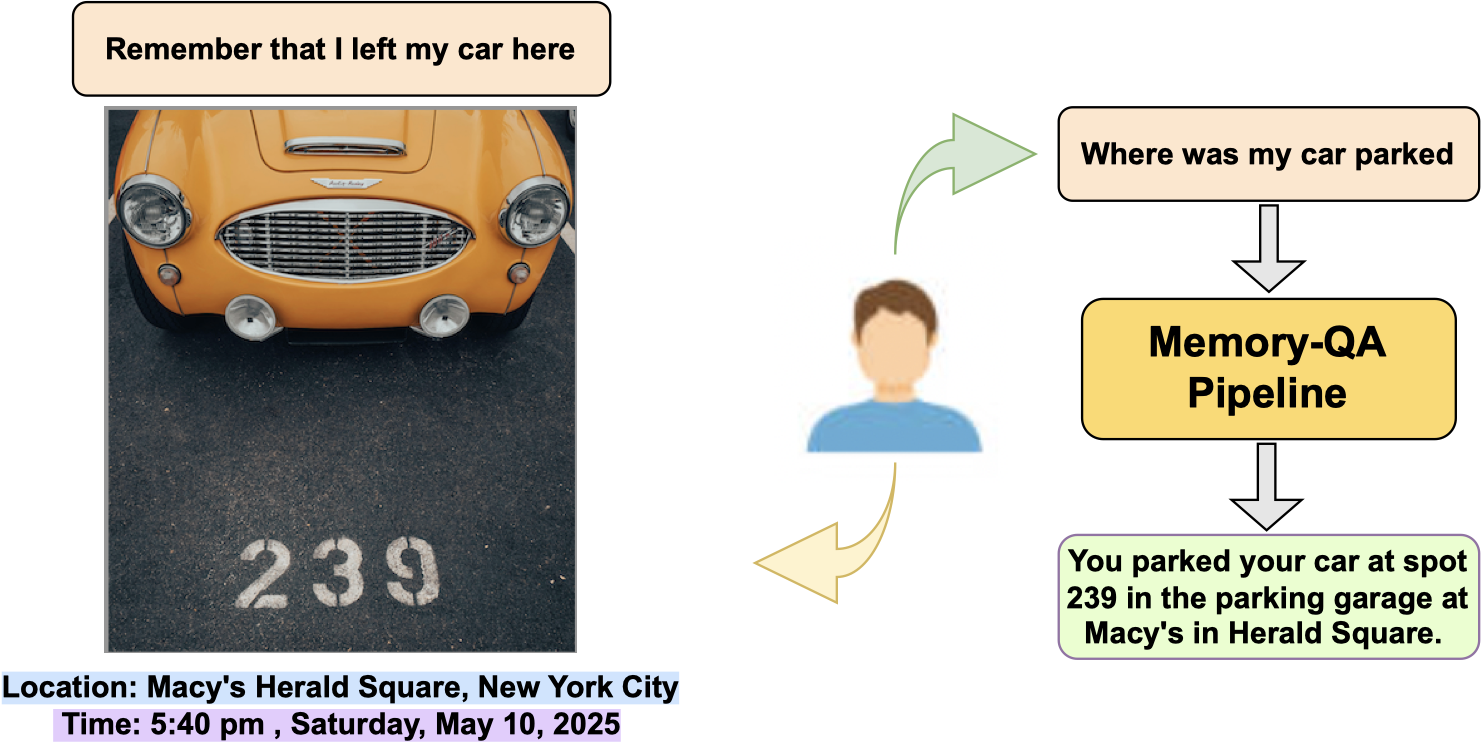}
\includegraphics[width=\linewidth]{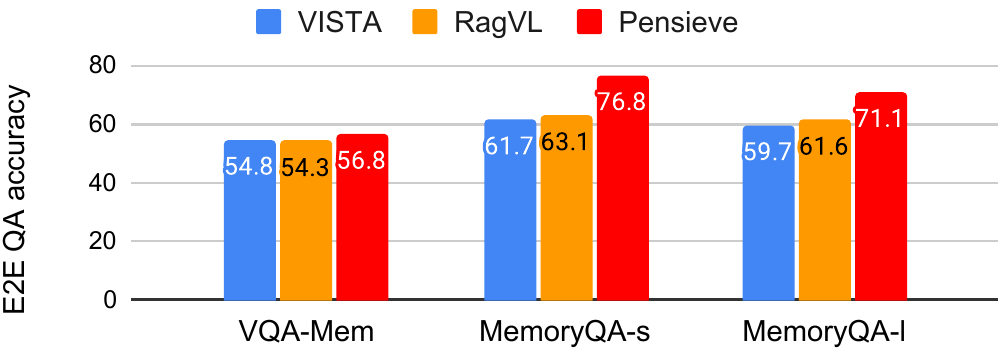}
\caption{A Memory-QA example. Our \system solution improves end-to-end QA accuracy over state-of-the-art MM-RAG systems~\cite{zhou2024vista, chen2024mllm} by up to 14\%.}
\label{fig:task}
\end{figure}

The rapid advancement in Vision-Language Models (VLM) has significantly enhanced the capabilities of intelligent assistants in understanding and reasoning over both textual and visual inputs, leading to substantial improvements in {\em Visual Question Answering (VQA)} \cite{yang2022empirical,zhang2023moqagpt}. Building on this progress, recent research on {\em Multi-Modal Retrieval-Augmented Generation (MM-RAG)} \cite{zhou2024vista,chen2024mllm} further extends these capabilities by first retrieving relevant images from large corpora, and then applying VQA techniques to the retrieved content.
At first glance, the Memory-QA task appears to align closely with MM-RAG. However, existing MM-RAG approaches still face several unique challenges when applied to Memory-QA scenarios.

To begin with, memory-related questions are often anchored to vague temporal or spatial references, such as {\em "yesterday"} or {\em "last month"} for time, and \textit{"at Macy's"} or \textit{"in downtown"} for location. Effectively leveraging such anchors is crucial for accurate question answering.
Moreover, many recall questions require aggregating information from multiple memory entries. For example, answering {\em "where did I park?"} typically involves retrieving the most recent parking memory, whereas {\em "what's on my shopping list?"} may require combining several past entries issued with {\em "remember to buy this"}.
Finally, most existing VLMs are constrained by limited visual context windows, which hinders their ability to reason over a large set of multimodal memory snapshots.

In this paper, we propose \system, the first end-to-end solution to the Memory-QA problem, grounded in three key intuitions. First, unlike standard MM-RAG tasks where the retrieval corpus is typically public and external, a personal memory repository resides in a personal context and can be explicitly augmented to enhance memory retention and retrieval. In the offline stage, we enrich each memory image with image captions generated by a Large Language Model (LLM), text extracted via Optical Character Recognition (OCR), and contextual metadata such as timestamps and geolocation. During memory retrieval, we propose a multi-signal retriever stack that incorporates temporal and location matching signals inferred from the user question. This dual-modality and condition-aware retrieval mechanism ensures more accurate and context-relevant memory selection.

Second, in contrast to general visual question types, such as object counting, spatial reasoning, activity recognition, and commonsense inference, recall questions tend to center around object and event tracking, which introduces opportunities for targeted optimization. To this end, we employ a few-shot learning image captioning module to better serve memory-related queries by predicting plausible recall questions for a given image, and incorporating the appropriate level of details in the generated captions to support such questions. Furthermore, we introduce a temporal query rewriter and fine-tune the answer generator at different stages to better align with the specific characteristics of Memory-QA.

Third, to enhance robustness and generalization, we employ multi-task instruction fine-tuning with noise injection, allowing the models to effectively handle the ambiguity and variability inherent in the retrieved memory candidates. At the question answering stage, we mitigate the challenge posed by limited context windows in VLMs by relying solely on the rich textual information generated during the memory augmentation phase. This design allows the system to reason across multiple memory snapshots without directly encoding raw images or large visual feature sets into the model's context.

In summary, our contributions are:
\begin{enumerate}
\vspace{-0.7 em}
    \item We formally define the Memory-QA problem, capturing key elements in real scenarios. We create a benchmark MemoryQA with 9,357 recall questions to illustrate real challenges.
\vspace{-0.7 em}
    \item We design the \system system for end-to-end memory-QA, improving quality with memory-specific augmentation, temporal-and-location-aware multi-signal retrieval, and fine-tuned multi-memory QA.
\vspace{-0.7 em}
    \item We conduct extensive experiments, showing \system improves over state-of-the-art MM-RAG solutions by up to 14\% on the MemoryQA benchmark. With \system, text-based LLMs obtain comparable results to VLMs, demonstrating a pathway to lower-cost Memory-QA.
\end{enumerate}

\section{Related Work}
\begin{figure*}[ht]
\centering
\includegraphics[width=\linewidth]{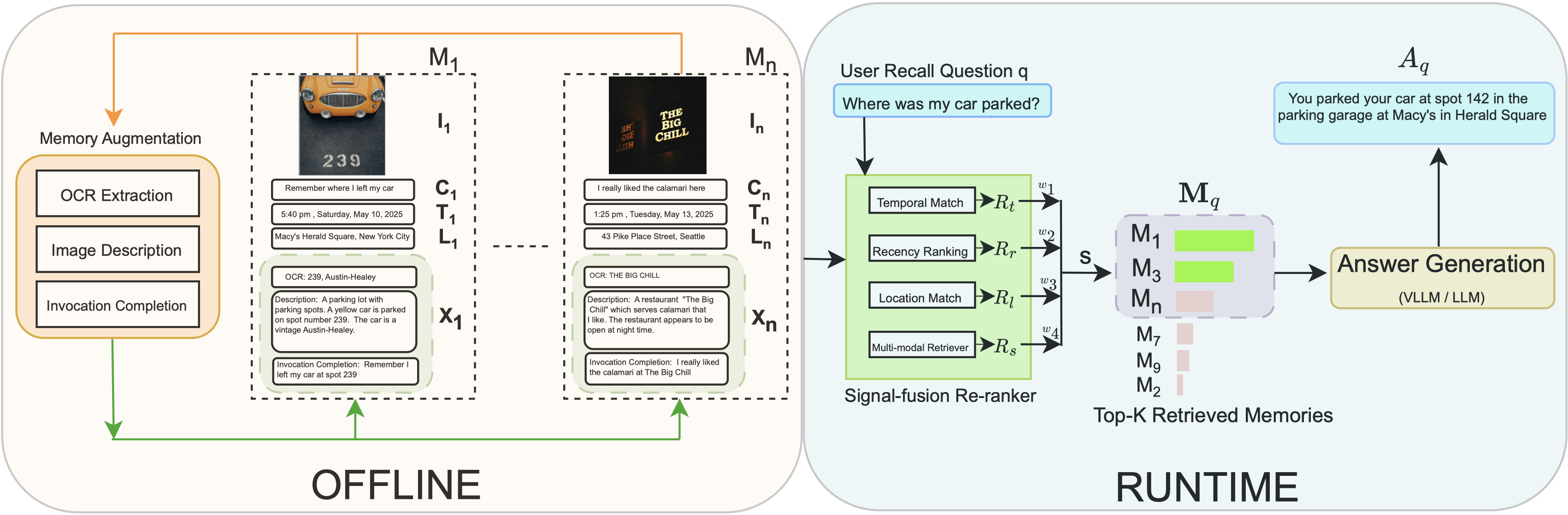}
\caption{Our proposed pipeline \system for Memory-QA.}
\label{fig:pipeline}
\end{figure*}

\paragraph{Multimodal Question Answering} requires reasoning on information from diverse modalities to answer questions.
It has evolved from early tasks focusing on vanilla VQA \citep{antol2015vqa}, to later expanding to more complex scenarios such as ManyModelQA \citep{hannan2020manymodalqa} and MultiModalQA \citep{talmor2021multimodalqa} which consider multiple modalities, as well as WebQA \citep{chang2022webqa} and SnapNTell \citep{qiu2024snapntell} involving real-world, knowledge-seeking questions.
To integrate these modalities, recent research has employed various approaches, including joint embedding of multiple modalities \citep{li2022mmcoqa,yu-etal-2023-unified}, fusing multimodal information with LLMs \citep{zhang2023moqagpt,yang2023enhancing,nan2024omg}, and transforming all modalities into text through captioning or description \citep{lin2022revive,yang2022empirical}.
Memory-QA differs from these works by focusing on recall questions to track events from previously created memories, introducing targeted optimizations.

\paragraph{Multimodal Retrieval-Augmented Generation} has been shown to enhance both the understanding and generation capabilities of vision models \citep{zheng2025retrieval}.
Pioneering works such as MuRag \citep{chen2022murag}, REACT \cite{liu2023learning} and RA-VQA \citep{lin2022retrieval} introduce language augmentations by parsing multimodal documents into text, aiming to improve retrieval and answer generation performance.
Recent studies have further refined this approach by enhancing model reasoning capabilities \citep{liu2023retrieval,tan2024retrieval} and robustness \citep{chen2024mllm,long2025retrieval}.
Another line of research has explored utilizing complete vision information by directly encoding images for retrieval and generating outputs solely based on visual content \cite{zhou2024vista,ma2024unifying,faysse2024colpali,yu2024visrag}.
However, these approaches fall short in addressing the temporal and location references inherent in Memory-QA.

\paragraph{Vision-Language Models} have been extensively explored in recent years to enhance fine-grained multimodal understanding. Building on the foundation laid by CLIP \citep{radford2021learning}, which enabled contrastive visual-text alignment, researchers have developed unified models that integrate pre-trained LLMs and vision encoders such as Flamingo \citep{alayrac2022flamingo} and BLIP-2 \citep{li2023blip}. Additionally, works like LLaVA \citep{liu2023visual}, mPLUG-Owl \citep{ye2023mplug}, and MiniGPT-4 \citep{zhu2023minigpt} have focused on fine-tuning LLMs for better visual feature alignment. More recently, models like Llama3.2 \citep{grattafiori2024llama} and Qwen2-VL \citep{wang2024qwen2} combine these techniques to achieve state-of-the-art open-source performances. These works have further enabled notable improvements in vision-language tasks including VQA \citep{hu2024bliva} and OCR \citep{shenoy2024lumos}. While some works have begun exploring multi-image understanding \citep{li2024llava}, handling large numbers of images or similar visual content remains challenging.

\section{Overview}
\label{sec:overview}

\subsection{Memory-QA Definition}
We now formally define the {\em Memory-QA} problem. Consider a repository of {\em memory entries} ${\cal M}=\{M_1, M_2, \dots, M_n\}$. Each memory entry is a tuple {$M_i = (I_i, C_i, T_i, L_i)$}, where $I_i$ denotes the image snapshot, $C_i$ denotes the invocation command for memory recording, $T_i$ denotes the timestamp of the memory, and $L_i$ denotes the location where the memory is captured. An invocation command, such as "remember this dress", often provides context information such as the reason for capturing the memory or the focus area in the view, critical for later recall; it is possible that the {\em invocation command} does not provide extra clues, such as a general command "remember this".

The {\em Memory-QA} problem takes as input a recall question $q$ asked at timestamp $T_q$ and generates an answer according to memories in $\cal M$. A good answer shall reflect information from all memories in $\cal M$ that are relevant to the input question $q$.

\subsection{Overview of \system}
As depicted in Figure~\ref{fig:pipeline}, \system consists of two parts: {\em offline augmentation} and {\em runtime QA}. Offline augmentation expands each memory entry with auxiliary text including OCR, image descriptions, and invocation completions that enrich the invocation commands (e.g., completing \textit{"remember where I parked"} with \textit{"remember I parked at slot 142"}).
After the augmentation the memory entry becomes $M_i^a = (I_i, C_i, T_i, L_i, X_i)$, where $X_i$ denotes the augmented auxiliary text.
Runtime QA takes user question $q$ and proceeds in two steps, similar to MM-RAG: first, it retrieves relevant memory candidates, denoted by ${\bf M}_q$; then, it generates the answer $A_q$ based on the retrieved memories.

Our solution incorporates three innovations. First, we introduce the task-oriented augmentation step for better memory retention. Second, we design memory-specific solutions for each step, including memory recording, memory retrieval, and answer generation, to optimize for Memory-QA. Third, our answer generation step is good at identifying and leveraging the set of memories that are necessary for answering the input question.

\section{Methodology}
\label{sec:method}

We now present \system in detail, highlighting how we leverage our three intuitions mentioned in Section~\ref{sec:intro} to address the challenges in Memory-QA.

\subsection{Memory Augmentation}
Offline memory augmentation takes as input a memory entry $M_i = (I_i, C_i, T_i, L_i)$, containing the snapshot image, the invocation command (text), the timestamp, and the location, and augments it with auxiliary memory clue $X_i$, which describes the memory snapshot in text. 

To facilitate subsequent runtime recall including retrieval and answer generation (\textbf{Intuition 1: Augmentation}), we create comprehensive memory clues by leveraging the invocation command $C_i$ and also predict potential memory-questions on the snapshot to generate the most effective memory clues (\textbf{Intuition 2: Memory-specific}).

\paragraph{Augmentation and encoding} Our auxiliary memory clue $X_i$ contains three fields: \textit{OCR results, image caption,} and \textit{invocation completion}, stored and indexed separately. First, since texts in the view are often critical in answering recall questions, such as {\em restaurant name, product price, name card,} and {\em phone number on a poster}, we apply OCR models to extract textual information from the memory image. Second, we leverage VLMs to generate detailed description of the image, to provide a foundation for understanding the visual content. Third, we complete the invocation command with the information in the image; taking \textit{"remember this restaurant"} as an example, the invocation completion can be "remember this Korean restaurant named Kochi". We do so by invoking VLMs to generate the complete command.

We use a multimodal encoder to embed the image and text concatenations. Formally,
\[
\Mmat_i = \Fcal(I_i, C_i, X_i, L_i)\in \Rcal^{d},
\]
where $\Fcal(\cdot)$ denotes the multimodal encoder, with $d$ being the embedding size.

\paragraph{QA-guided Image Description Generation}
Given that Memory-QA focuses on task-oriented questions with the intention of recalling useful information, we can enhance vanilla image captioning to generate descriptions specifically effective in answering a wide range of memory recall questions posed on the memory entry.
The QA-guided image description generation is achieved through a few-shot learning approach using VLMs.

Starting with example task-oriented questions, we prompt a VLM to generate potential recall questions on the image and the answers to the questions; we use in-context learning and provide a diverse list of recall questions as few-shot examples. We then prompt the model to generate a comprehensive image description that enables answering the recall questions without referring to the image. 
The resulting model will be able to focus on salient, question-relevant image features for captioning.
The generated QA-guided image descriptions are used as a crucial memory element for retrieval and answer generation.

\subsection{Memory Retrieval}
The runtime multimodal retrieval step takes as input a recall question $q$ and the augmented memory repository ${\cal M}^a$, identifies a set of memory entries ${\cal M}^a_q$  that are relevant to the question. 

To effectively utilize all memory components, we employ a multi-signal retriever that integrates information from snapshot images, invocation commands, temporal/location context, and offline-generated augmentations. Our retrieval stack first runs a temporal and location matching module in parallel with a multimodal retriever. The matching module computes date match scores, location match scores, and recency scores, while the multimodal retriever predicts similarity scores. These independent signals are then combined by a re-ranker to produce the final ranked list of memories.

\paragraph{Temporal / location matching module} 
We first employ an LLM-based date parser to extract temporal information from the raw query, specifically the search date range $(T_s, T_e)$ and a boolean $B_{r}$ indicating whether recent memories should be favored. This extracted information is used to calculate date match and recency scores as follows.

For queries with a non-empty search date range (e.g., \textit{``What did I save last week?''}), the date match score $R_t$ for memory $M_i$ is computed as:
\[
R_t(M_i, q, T_q) = {\bm 1}\{T_s(q, T_q) \le T_i \le T_e(q, T_q)\},
\]
where $\mathbf{1}{\cdot}$ denotes an indicator function.

For queries seeking recent memories (e.g., \textit{``Where did I park last time?”}), the recency score $R_r$ for memory $M_i$ is calculated as:
\[
R_r(M_i, T_q) = B_{r}\Big(e^{-\frac{\delta}{Q_{\rm s}}} + e^{-\frac{\delta}{Q_{\rm m}}} + e^{-\frac{\delta}{ Q_{\rm l}}}\Big)/3,
\]
where $\delta = T_q - T_i$ represents the interval between memory creation and the query time. The constants $Q_{\rm s} = 3$ days, $Q_{\rm m} = 90$ days, $Q_{\rm l} = 365$ days, simulate varying rates of memory decay over short/middle/long time periods \citep{li2023tradinggpt}. Note that if the parser predicts $B_{r}=0$, all recency scores default to zero.

To prioritize memories whose creation locations match the user’s query, we compute a location match score $R_l$ using the BM25 algorithm:
\[
R_l(M_i, q) = {\rm BM25}(L_i, q).
\]
\paragraph{Multimodal retriever} We use a multimodal encoder to embed both memory content and the query. The similarity score $R_s$ is defined as the dot product between these embeddings:
\[
R_s(M_i^a, q) =\Mmat_i^{\top}\cdot\Fcal(q),
\]
where $\Fcal(\cdot)$ represents the multimodal encoder. $\Mmat_i=\Fcal(M_i^a)$ is the memory embedding produced by the same encoder during offline encoding.

\paragraph{Signal fusion re-ranker:} Our re-ranker integrates the temporal/location matching signals and the multimodal retriever signal. For a given query $q$, the final retrieval score $s_i$ for memory candidate $M_i$ is computed as a weighted sum:
\[
\begin{aligned}
s_i = &\, w_tR_t(M_i, q, T_q) + w_rR_r(M_i, T_q) \\
& + w_lR_l(M_i, q) + w_sR_s(M_i^a, q),
\end{aligned}
\]
where $w_t$, $w_r$, $w_l$, and $w_s$ are weights for each signal. To enable domain-specific customization, we optimize the retriever weights by training a linear model on a small volume of domain data (see Appendix \ref{sec:reranker}).
This approach allows us to adapt the model to the specific memory domain while providing additional model interpretability through the learned weights. In the end, we send the top-K candidates for answer generation: $
{\cal M}^a_q = {\rm TopK}\ \{s({\cal M}^a, q)\}.
$

\subsection{Answer Generation}
The runtime answer generation step takes as input a recall question $q$ and the retrieved memory set ${\cal M}^a_q \subset{\cal M}^a$, and generates the answer $A_q$ to the question, $A_q = Gen(q, {\cal M}^a_q)$.
In this work, leveraging the high-quality memory augmentations, we argue that only using the rich textual memories to generate $A_q$ with a text-based LLM is not only lower-cost, but also achieves comparable performances as using multimodal memories.

The main challenge for this step is that answer generation may need to aggregate information from multiple sources. We conduct fine-tuning such that the VQA model can effectively identify positive and negative candidates from ${\bf M}_q$ and answer the question only based on relevant memories (\textbf{Intuition 3: Multi-memory QA}) .

\paragraph{Noise-injected instruction tuning}
To mitigate the negative impact of irrelevant memories retrieved, we employ noise-injected training \citep{chen2024mllm}.
This approach creates the training dataset by including up to 2 confusing candidates as negative examples, which are presented alongside positive memories in a similar manner as mentioned above.
By doing so, the fine-tuned model is trained to robustly discern relevant from irrelevant information, thereby strengthening its memory comprehension and ability to filter out noise.

\paragraph{Multi-task instruction tuning}
Considering the nature of this answer generator is to do two tasks at the same time: detect positive candidates from ${\bf M}_q$ and generate an answer $A_q$ to the question based on relevant candidates, we propose multi-task fine-tuning to jointly train two tasks simultaneously and improve the answer correspondence with relevant memories.
Specifically, the LLM outputs {\em a list of positive memory Ids followed by the generated answer}. Training aims to optimize a standard autoregressive cross-entropy loss $L = -\sum_{t=1}^T\log P(y_t|y_{<t})$ computed over the entire response, where $T$ is the total sequence length.

\section{Experiment Setup}
\label{sec:exp_setup}

\begin{table}[t!]
    \centering
\resizebox{\columnwidth}{!}{%
    \begin{tabular}{lrrcc}
       \toprule
       Datasets & \#Images & \#Samples & \makecell{Recall Question\\Only} & \makecell{Time \&\\Location} \\
       \midrule
       \multicolumn{5}{l}{MemoryQA} \\
       \ \ train & 3,011  & 6,386   & Yes & Yes \\
       \ \ test-$s$ & 189  & 1,326   & Yes & Yes \\
       \ \ test-$l$ & 2,789  & 2,971   & Yes & Yes \\
       \midrule
       VQA-Mem & 1,469 & 1,811 & Yes & No \\
       WebQA & 39,000    & 2,511  & No & No \\
       \bottomrule
    \end{tabular}
    }
    \caption{Statistics of the datasets used in this work.}
    \label{tab:datasets}
\end{table}
\subsection{Datasets}
We experiment with three benchmarks: our in-house dataset MemoryQA\footnote{MemoryQA dataset will be available at: \url{https://github.com/facebookresearch/MemoryQA/}}, an extended version of VQA~\citep{antol2015vqa}, called VQA-Mem, and WebQA~\citep{chang2022webqa} (statistics shown in Table~\ref{tab:datasets}).
All datasets are formatted as positive and negative samples for each question, with VQA-Mem's negative samples being randomly selected to synthesize user memory diversity.

\paragraph{MemoryQA}
Our in-house benchmark, MemoryQA, comprises 5,800 images captured from daily life using wearable devices such as smart glasses. What sets MemoryQA apart from other multimodal QA benchmarks is the inclusion of temporal and location information for each image. For the test sets, we employ human annotators to craft invocation commands and timestamped recall questions for each image, and also to verify the accuracy of each answer (see Appendix \ref{sec:mem-QA-dataset} for more details). In contrast, the training set was generated using VLM without human annotations. We have two test sets with different sizes, MemoryQA-$s$, and MemoryQA-$l$ with more challenging questions and more diverse image types.

\paragraph{VQA-Mem}
To adapt VQA for the Memory-QA task, we first prompt Llama3.3-70B to remove QA pairs that do not pertain to recall questions and answers, and only keep images with at least one recall QA pair, ensuring relevance for Memory-QA.
Then, for each selected image, we use Llama3.2-90B to generate two different invocation commands that trigger memory recording.

\paragraph{WebQA}
We utilize the image modality data from the WebQA validation set.
We keep the original images and QA pairs without filtering, so this is more of a standard multimodal QA dataset.
We treat image titles as invocation commands for our problem setting.

\subsection{Evaluation}
Our overall metric is the \textit{E2E QA accuracy} for the recall questions, computed as the percentage of questions that are correctly and completely answered. We compare an answer with the ground truth and decide its correctness in three ways: computing the keyword overlaps~\citep{chang2022webqa}, LLM-as-a-judge \citep{li2024generation} with Llama3.3-70B, and decide if the ground truth is entailed in the generated answer~\citep{lattimer2023fast}. We denote the accuracy computed in these metrics by $A_{key}$, $A_{llm}$, and $A_{ent}$ respectively.

\begin{table}[t!]
    \centering
    \small
    \resizebox{\columnwidth}{!}{%
    \begin{tabular}{lcccc}
       \toprule
       Model & WebQA & VQA-Mem & \multicolumn{2}{c}{MemoryQA} \\
        &  &  & $s$ & $l$ \\
       \midrule
       \multicolumn{5}{c}{\textit{Baseline (w/o aug., CLIP)}} \\
       GPT-4o (vis) & 64.4 & 54.2  & 58.2 & 49.2 \\
       Llama3.2-90B (vis) & 54.0  & 50.9  & 53.1 & 46.9 \\
       Llama3.3-70B (txt) & 13.8 & 6.5  & 27.0 & 28.4 \\
       \midrule
       \multicolumn{5}{c}{\textit{SOTA MM-RAG systems, w/ GPT-4o}} \\
       VISTA &  69.1 & 54.8 & 61.7 & 59.7 \\
       RagVL & \textbf{71.1}  & 54.3 &  63.1 & 61.6 \\
       \midrule
       \multicolumn{5}{c}{\textit{\system: w/ aug., Multi-signal Retriever}} \\
       GPT-4o (vis) & 66.4  & \textbf{56.8} &  \textbf{76.8} & \textbf{71.1} \\
       Llama3.2-90B (vis) & 59.4  & 53.8  & 74.7 & 68.5 \\
       Llama3.3-70B (txt) & 39.9  & 46.9 & 74.1  & 70.3 \\
       \bottomrule
    \end{tabular}
    }
    \caption{E2E QA results $A_{llm}$. \system outperforms the baseline and state-of-the-art solutions on recall questions from VQA-Mem and MemoryQA by a big margin.}
    \label{tab:overall}
\end{table}

\begin{table}[t!]
    \centering
    \small
    \resizebox{\columnwidth}{!}{%
    \begin{tabular}{lcccc}
       \toprule
       Method & \multicolumn{2}{c}{VQA-Mem} & \multicolumn{2}{c}{MemoryQA-$s$} \\
       \midrule
       \textit{Vision Methods} & $A_{key}$ & $A_{llm}$ & $A_{key}$ & $A_{llm}$ \\
       GPT-4o & \textbf{55.8} & \textbf{56.8} & \textbf{69.4} & \textbf{76.8} \\
       \quad w/o augmentation & 52.6 & 53.9  & 61.6 & 67.9 \\
       \quad w/o MS retriever & 53.2 & 53.8 & 59.8 & 65.9 \\
       \quad w/o QA-guided desc. & 53.4 & 55.2 & 67.9 & 75.6  \\
       \quad w/o time/loc. match & - & - & 64.0 & 69.8 \\
       \midrule
       \textit{Text Methods} & $A_{key}$ & $A_{llm}$ & $A_{key}$ & $A_{llm}$ \\
        Llama3.3-70B & \textbf{47.5} & \textbf{46.9}  & \textbf{71.0} & \textbf{74.1}\\
        \quad w/o augmentation & 7.3 & 6.4 & 38.5 & 31.7 \\
       \quad w/o MS retriever & 47.0 & 46.0 & 61.7 & 63.1 \\
       \quad w/o QA-guided desc. & 42.7 & 43.1 & 70.5 & 71.0  \\
       \quad w/o time/loc. match & - & - & 66.9 & 67.9 \\
       \bottomrule
    \end{tabular}
    }
    \caption{Ablation study on both datasets shows significant improvements from our design choices. }
    \label{tab:ablation}
\end{table}

\begin{table*}[t!]
\centering
\small
\begin{tabular}{lccccccccccc}
\toprule
Retriever & Reranker & \multicolumn{2}{c}{R@1} & \multicolumn{2}{c}{R@3} & \multicolumn{2}{c}{R@5} & \multicolumn{2}{c}{nDCG@3} & \multicolumn{2}{c}{nDCG@5} \\
\cmidrule(lr){3-4}
\cmidrule(lr){5-6}
\cmidrule(lr){7-8}
\cmidrule(lr){9-10}
\cmidrule(lr){11-12}
 &  & $s$ & $l$ & $s$ & $l$ & $s$ & $l$ & $s$ & $l$ &  $s$ & $l$  \\
\hline
\multicolumn{12}{c}{\textit{Baseline retriever}} \\
BM25 & - & 11.2 &  11.1 & 21.4 & 23.3  & 26.8 & 31.4  & 17.1 & 18.8  & 19.3  & 22.2 \\
Dragon+ & - & 69.6  & 71.0 & 82.2  & 82.5 & 85.8  & 86.0 &  77.1 & 82.2  & 78.5  & 83.5 \\
CLIP  & - & 66.2 & 59.3 & 79.4 & 72.8  & 84.4 & 77.1  & 73.8 & 70.3  & 75.9  & 72.0 \\
Vis-BGE-base & - & 69.2  & 64.8 & 82.2  & 76.2 & 85.8  & 80.1 & 76.9  & 75.0 & 78.4  & 76.5 \\
Vis-BGE-m3 & - & 73.5  & 71.2 & 85.5  & 82.7 & 88.4  & 86.4 &  80.7  & 82.2 & 81.9  & 83.6 \\
RagVL & VLM & 71.4 & 74.2 & 85.8 & 85.8  & 88.4 & 88.3  & 80.4  & 85.0  & 81.6  & 85.9 \\ 
\hline
\multicolumn{12}{c}{\textit{Vis-BGE-m3 + temporal/location matching module + Reranker}} \\
Multi-signal & Max & 71.9 & 71.7  & 85.2 & 88.2   & 89.8 & 91.7  & 79.7 & 84.4 & 81.6  & 85.8 \\
Multi-signal &  Sum & 80.6 & 77.9  & 91.8 & 91.3  & 94.2  & 94.2  & 87.3 & 89.2  & 88.3   & 90.7\\
Multi-signal & Learned weights & \textbf{84.3} & \textbf{80.9}  & \textbf{93.5} & \textbf{92.1}  & \textbf{95.5}  & \textbf{95.0}  & \textbf{89.8} & \textbf{91.0}  & \textbf{90.6}  & \textbf{92.0} \\
\hline
\multicolumn{12}{c}{\textit{Ablation study}} \\
w/o date match score & Learned weights & 76.2 & 73.4  & 86.8 & 83.7  & 89.4 & 87.3 & 82.5 & 83.9  & 83.6  & 85.2 \\
w/o recency score & Learned weights & 82.6 & 79.1  & 92.8 & 91.0  & 94.8 & 94.3  & 88.7 & 89.6 & 89.6  & 90.7 \\
w/o location score & Learned weights & 83.5 & 80.3 & 93.1 & \textbf{92.1}  & 95.3 & 94.9  & 89.4 & 90.7  & 90.4  & 91.7 \\
\bottomrule
\end{tabular}
\caption{Performance comparison of different retriever and reranker configurations on MemoryQA-$s$ and MemoryQA-$l$. Each question has 10 to 50 candidate memories.}
\label{tab:retriever_reranker}
\end{table*}

\subsection{Implementation}
In the implementation of \system, we use techniques introduced in Section \ref{sec:method}.
Notably, all memory augmentations are generated once before experimentation using Lumos \citep{shenoy2024lumos} for OCR and Llama3.2-90B for image descriptions and invocation completions.
We compare \system with a baseline solution that records memories without augmentations, relying on retrieval with CLIP \citep{radford2021learning} or Dragon+ \citep{lin2023train} embedding similarity for image retrieval. 
We also include state-of-the-art MM-RAG systems VISTA \citep{zhou2024vista} and RagVL \citep{chen2024mllm} for comparison.

\section{Results}
\label{sec:experiments}

We now present comprehensive experimental results to show the performance of our \system system, and justify our various design choices. 

\subsection{Overall results}
Table~\ref{tab:overall} compares our \system solution with baselines and state-of-the-art systems.
We have four observations. First, our approach outperforms the current SOTA MM-RAG methods by a significant margin (+14\%) on MemoryQA-$s$ and (+10\%) on MemoryQA-$l$, and also improves on VQA-Mem (+2\%), highlighting the effectiveness of our targeted solutions for this task. Although WebQA lacks recall questions, our memory-based adaptation only slightly regress the results. Second, surprisingly, even using a text LLM (Llama3.3-70B) for answer generation, we achieve comparable QA accuracies over VLMs on MemoryQA, showing a cost-effective approach for Memory-QA. Third, we observe that across different backbone models, our solutions consistently outperform vision baselines by 19\% on MemoryQA-$s$ and 22\% on MemoryQA-$l$. Notably, the even larger improvements (>42\%) are observed with text models because of the textual augmentations. Finally, despite MemoryQA-$l$ being a much more challenging and diverse dataset, \system still achieves robust performances with 71\% E2E QA accuracy.

\begin{table}[t!]
    \centering
    \small
    \begin{tabular}{lrrrr}
       \toprule
       Memory & sim-p & sim-n & diff & $A_{key}$ \\
       \midrule
       \multicolumn{5}{c}{\textit{Accumulative Text - Llama3.3-70B}} \\
       Invocation command & 32.4 & 14.2 & 18.2 & 7.3 \\
        + OCR result & 38.9 & 17.5 & 21.4 & 23.5 \\
        + Inv. completion & 44.4 & 19.4 & 25.0 & 31.4 \\
       + Image caption & 53.4 & 26.0 & 27.4 & 42.7 \\
       + QA-guided caption & 55.1 & 24.5 & 30.6 & 47.5 \\
       \bottomrule
    \end{tabular}
    \caption{Memory augmentation increases embedding differences between positive and negative answers, thus increases QA quality. }
    \label{tab:memory}
\end{table}

\subsection{Ablation Study}
Table~\ref{tab:ablation} shows a system ablation study on MemoryQA-$s$ and VQA-Mem, which focus on recall questions. 
In sum, removing any component results in lower E2E QA results on both evaluation metrics.
We observe significant improvements brought by our major design choices. Taking vision methods and the MemoryQA-$s$ dataset as an example, (1) memory augmentation improves $A_{llm}$ by 9\%; (2) multi-signal retrieval leveraging the invocation commands, raw information about time and location, and the augmentations improves $A_{llm}$ by 11\%; (3) QA-guided description generation improves $A_{llm}$ by 1\%; and (4) time/location matching improves $A_{llm}$ by 7\%. These improvements are even more pronounced when we use text-based models, where the extra signals play an important role when vision references are absent.

\subsection{Memory Augmentation Results}
To investigate the impact of each memory augmentation component, we conduct experiments on VQA-Mem. In addition to end-to-end QA accuracy measured by $A_{key}$, we show embedding similarity scores between the textual memory components and the QA pairs, where higher scores for positive samples and lower scores for negative samples indicate more effective augmentation. Table~\ref{tab:memory} shows the metrics as we progressively add OCR results, invocation completions, image captions, and QA-guided captioning. As we add more augmentations, we observe the gap between embedding similarity with positive samples and negative samples gets bigger, leading to higher QA accuracy.

\begin{table}[t!]
    \centering
    \small
    \begin{tabular}{lccc}
       \toprule
       Answer generator & $A_{key}$ & $A_{llm}$ & $A_{ent}$ \\
       \midrule
       GPT-4o (\textit{vision}) & 69.4  & 76.8  & 64.4 \\
       Llama3.2-90B (\textit{vision}) & 70.0  & 74.7 & 62.1 \\
       Llama3.3-70B (\textit{text}) & 71.0 & 74.1 & 61.1 \\
       \midrule
       Llama3.1-8B (\textit{text})  &  65.2 & 66.9 & 56.2 \\
      Llama3.1-8B-SFT (\textit{text})  & 68.6  & 72.4 & 61.3 \\
      \quad  w/o noise-injection  & 67.5  & 69.6 & 59.4 \\
      \quad  w/o multi-tasking  & 68.4  & 70.4 & 60.4 \\
       \bottomrule
    \end{tabular}
    \caption{Answer generation performances, highlighting similar performance between vision and text models, and effectiveness of fine-tuning.}
    \label{tab:answer}
\end{table}

\begin{table}[t!]
    \centering
    \small
    \begin{tabular}{lccc}
       \toprule
       Answer generator & precision & recall & F1 \\
       \midrule
       GPT-4o (\textit{vision}) & 83.5  & \textbf{93.4}  & 88.2  \\
       Llama3.2-90B (\textit{vision}) & 85.0 & 90.2 & 87.5 \\
       Llama3.3-70B (\textit{text}) & 86.0  & 92.6 & 89.2 \\
       Llama3.1-8B (\textit{text})  & 85.6 & 90.8 & 88.1  \\
       Llama3.1-8B-SFT (\textit{text})  & \textbf{87.8} & 91.2  & \textbf{89.5}  \\
       \bottomrule
    \end{tabular}
    \caption{Guided decoding performances to detect positive candidates during answer generation.}
    \label{tab:coded}
\end{table}

\subsection{Multimodal Retrieval Results}
We evaluate the effect of multi-signal retriever on MemoryQA, the only dataset that includes temporal and location references. We use recall@$k$ and nDCG@$k$ to measure how well the correct memories are retrieved and ranked among the top-$k$ candidates.
In Table \ref{tab:retriever_reranker}, we compare our methods with a set of alternatives.
The retrieval performances on MemoryQA-$s$ and MemoryQA-$l$ generally follow similar trends.
First, we test a set of text-based and visual-based baselines, showing that visual-BGE-m3 models~\citep{zhou2024vista} and RagVL with an VLM reranker~\citep{chen2024mllm} achieve the best baseline performances by leveraging both textual and visual information; however, our multi-signal retriever integrates similarity with temporal-location relevancy signals, improving various metrics by up to 13\%.
Second, we compare different ways to combine the various signals. In particular, using learned weights achieves a Recall@5 of 0.955 on MemoryQA-$s$ and 0.950 on MemoryQA-$l$.
The learned weights here are $0.08$, $0.22$, $0.16$ and $0.53$ for $w_t$, $w_r$, $w_l$ and $w_s$ respectively. This suggests that while the multimodal retrieving signal ($w_s$) dominates with half of the weights, the temporal ($w_r$) and location ($w_l$) signals takes the rest half, playing crucial roles in memory retrieval.
Finally, we compared with removing the time and location signals, finding that omitting date-matching signals has the most substantial impact on performance.

\begin{figure}[t!]
    \centering
    \includegraphics[width=\linewidth]{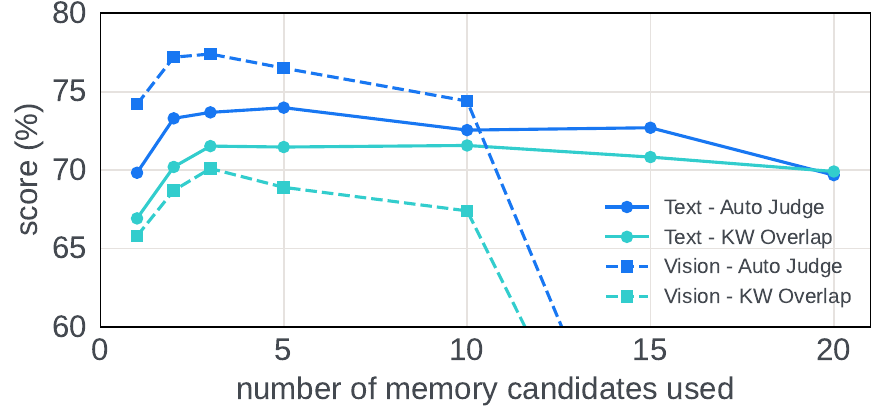}
    \caption{Generated answer quality vs number of memory candidates in the answer generator prompt. }
    \label{fig:scores_vs_topk}
\end{figure}

\begin{table*}[t!]
    \centering
    \small
    \begin{tabular}{lccc}
       \toprule
       Error Bucket & Baseline & \system-vision & \system-text \\
       \midrule
        Correct & 772 (58.2\%)  & 	1018 (76.8\%)  & 982 (74.1\%) \\
        Retrieval error: not all positive memories are retrieved & 256 (19.3\%) & 83 (6.3\%) & 82 (6.2\%) \\
        Generation error: response from wrong memories & 50 (3.8\%) & 47 (3.5\%) & 49 (3.7\%) \\
        Generation error: missing key info in augmentations & N/A & 90 (6.8\%) & 123 (9.3\%) \\
        Generation error: temporal reasoning, LLM-Judge, etc. & 248 (18.7\%) & 88 (6.6\%) & 90 (6.8\%) \\
       \bottomrule
    \end{tabular}
    \caption{Error analysis on MemoryQA-$s$ with categorized failure cases. GPT-4o backbones Baseline and \system-vision, and Llama3.3-70B backbones \system-text. Augmentation errors are not applicable for baselines.}
    \label{tab:error}
\end{table*}

\subsection{Answer Generation Results}
Finally, we evaluate the performance of answer generation and verify the effectiveness of our fine-tuned model on MemoryQA-$s$. To ensure controlled comparisons, all models are fed with the same top 3 retrieved candidates. In addition to QA accuracy (Table~\ref{tab:answer}), we also compute the precision and recall of the answer generator in identifying positive candidates (Table~\ref{tab:coded}).
The comparisons confirm that relying solely on textual memory augmentations is an equally effective alternative to utilizing visual content for MemoryQA. In addition, it shows that fine-tuning Llama3.1-8B yields comparable QA results to 70B and achieves the best F1 score in detecting positive candidates, despite being a smaller model. Finally, omitting noise injection or multi-task instruction tuning significantly impacts the fine-tuned model's performance.

Figure~\ref{fig:scores_vs_topk} plots answer accuracy versus the number of retrieved candidates. Text methods get the highest accuracy with 5 retrieval candidates, and then the accuracy flattens out till reaching 20 results, showing the robustness of textual augmentations. In contrast, vision methods get highest accuracy with 3 candidates, suffer significantly when the number of memory candidates increases, because of the large visual context size.

\begin{table}[t!]
    \centering
    \small
    \begin{tabular}{lccc}
       \toprule
Component & Config & P50 (ms) & P90 (ms) \\
\midrule
\multicolumn{4}{l}{\textit{E2E}} \\	
Runtime & w/ API (text) & 1400 & 1950 \\
Offline & default & 1800 & 4710 \\
\midrule
\multicolumn{4}{l}{\textit{Runtime components}} \\	
Retrieval & local \& API & 630 & 880 \\
\ \ Query Embed & local GPU & 18 & 19 \\
\ \ Datetime match & API & 600 & 850 \\
Generate answer & API (text) & 630 & 920 \\
Generate answer & API (vision) & 740 & 1470 \\
\makecell{Generate answer\\(fine-tuned 8B)} & local GPU & 920 & 1200 \\
\midrule
\multicolumn{4}{l}{\textit{Offline components}} \\	
OCR & API & 500 & 1200 \\
Image caption & API & 1500 & 4350 \\
Memory Embed & local GPU & 190 & 260 \\
       \bottomrule
    \end{tabular}
    \caption{Latency analysis on the proposed system. P50 and P90 denote 50th and 90th percentile latency. Here we use 40GB A100 GPU as the local GPU.}
    \label{tab:latency}
\end{table}

\subsection{Error Analysis}
To inform future improvements, we performed a detailed analysis of model errors and categorized failure cases, as shown in Table \ref{tab:error}.
\system substantially reduced both retrieval errors and temporal reasoning errors compared to the baseline (from 19\% to 6\%). The most frequent error in \system-text is “missing key info in memories” (9.3\%), which likely stems from image captioning omitting important details. In contrast, \system-vision exhibits fewer such errors (6.8\%).

\subsection{Latency Analysis}
Table \ref{tab:latency} shows the latency analysis. The total estimated p50 latency for runtime QA is 1.4 seconds, which enables productionization in most AI assistants. During runtime, answer generation is the primary contributor to latency, with the vision-based approach taking 50\% longer than the text-based approach at p90. This result highlights the advantage of \system-text when its performance is competitive. In the offline stage, latency requirements are normally less stringent, and the main source of delay is image captioning.

\section{Conclusion}
\label{sec:conclusion}

In this paper, we introduce the Memory-QA task as a critical step towards realizing the long-standing vision of a second brain. We propose \system, a novel end-to-end Memory-QA system integrating multimodal memory recording, memory retrieval, and answer generation. To address the challenges in Memory-QA, we propose targeted solutions including memory-specific augmentations, QA-guided image description generation, temporal and location-fused multi-signal retrieval, and multi-memory QA fine-tunings. To facilitate research in this area, we create a new multimodal QA dataset and extend existing VQA benchmarks with memory-centric recall questions tailored to this new task. We conduct extensive experiments, demonstrating the effectiveness, efficiency, and adaptability of our approach on Memory-QA.

\section{Limitations}
Despite the contributions of this research, there are several limitations that warrant consideration. Firstly, while the proposed approach is applicable to other multimodal QA tasks, its performance on these tasks may not be guaranteed to match the level achieved on the Memory-QA task. Secondly, to ensure fair comparisons with API-based models such as GPT-4o, we did not fine-tune large-scale LLMs or VLMs, which could have further improved the models' performances. Finally, the datetime matching module in the proposed multi-signal retriever has limitations in handling specific holidays whose dates change annually, which could impact its accuracy in certain scenarios.

\section{Potential Risks and Ethical Considerations}
Our work adheres to high ethical standards in data collection, usage, and publication. Below, we outline key aspects of our ethical compliance to potential risks:
\begin{itemize}
    \item The dataset used in this study contains no offensive, harmful, or inappropriate content.
    \item Any PII, such as physical addresses, was synthetically generated and does not correspond to real individuals. 
    \item All images in the Memory-QA dataset were collected with informed consent from contributors and are approved for research use. Human faces were blurred and manually verified, and any images containing unblurred or clearly identifiable faces were removed.
    \item The dataset does not contain, infer, or annotate any protected or sensitive attributes, such as sexual orientation, political affiliation, religious belief, or health status.
    \item No characteristics of the human subjects (e.g., age, gender, identity) were self-reported or inferred during data collection or analysis.
    \item The dataset is made available strictly for research purposes. It is not intended for commercial use or deployment in any application.
\end{itemize}

\bibliography{custom}

\appendix

\section{Appendix}
\label{sec:appendix}

\subsection{Answer Generation Evaluation Metrics}
\subsubsection{Keywords Overlapping Accuracy}
Keywords overlapping accuracy is proposed by \citet{chang2022webqa} that aims to:
1. Detect the presence of key entities. 2. Penalize the use of any incorrect entities. 3. Avoid penalizing semantically relevant but superfluous words.
The answer domains $D_{qc}$ are defined for the question categories $qc$ including color, shape, number, Y/N questions, and others.
With $c$ as a candidate output, $K$ for correct answer keywords, and $qc$ for question category, the keywords overlapping accuracy is computed by
\[
A_{\rm key}(c, K) = 
\begin{cases} 
    \displaystyle F_1 \left( c \cap D_{qc}, \, K \cap D_{qc} \right) \\ \quad {\small\text{if } qc \in \{\text{color, shape, number, Y/N}\}} \\
    \displaystyle RE \left( c, \, K \right) \\\quad {\small\text{otherwise}},
\end{cases}
\]
where $RE$ denotes the recall BARTScore.

\subsubsection{Auto Judge}
We use an LLM-based (Llama3.3-70B) auto judge to compare the generated answers with the ground truth answers. The prompt for auto judge is shown in Figure \ref{fig:autojudge}.

\subsubsection{Entailment Score}
For the entailment score, we use SCALE \citep{lattimer2023fast}, an automatic evaluation method that provides general faithfulness and factuality scores on all text generations. The generated answer is fed into a prompt to detect whether it implies the ground truth answer.
Each prompt is then run through Flan-T5 \citep{chung2024scaling}, a pre-trained sequence-to-sequence LLM, and the resulting logits are used to compute the entailment scores.
Specifically, logits are obtained by prompting $M$ with the following: \textit{$l$ = $M$(“\{premise\} Question: does this imply
‘\{hypothesis\}’? Yes or no?”)}. The entailment probability is then calculated by
\[
P_{\rm entail} = {\text{SoftMax}}(l[{\rm ``Yes"}], l[{\rm ``No"}])[0].
\]

\begin{figure}[t!]
\centering
\includegraphics[width=\linewidth]{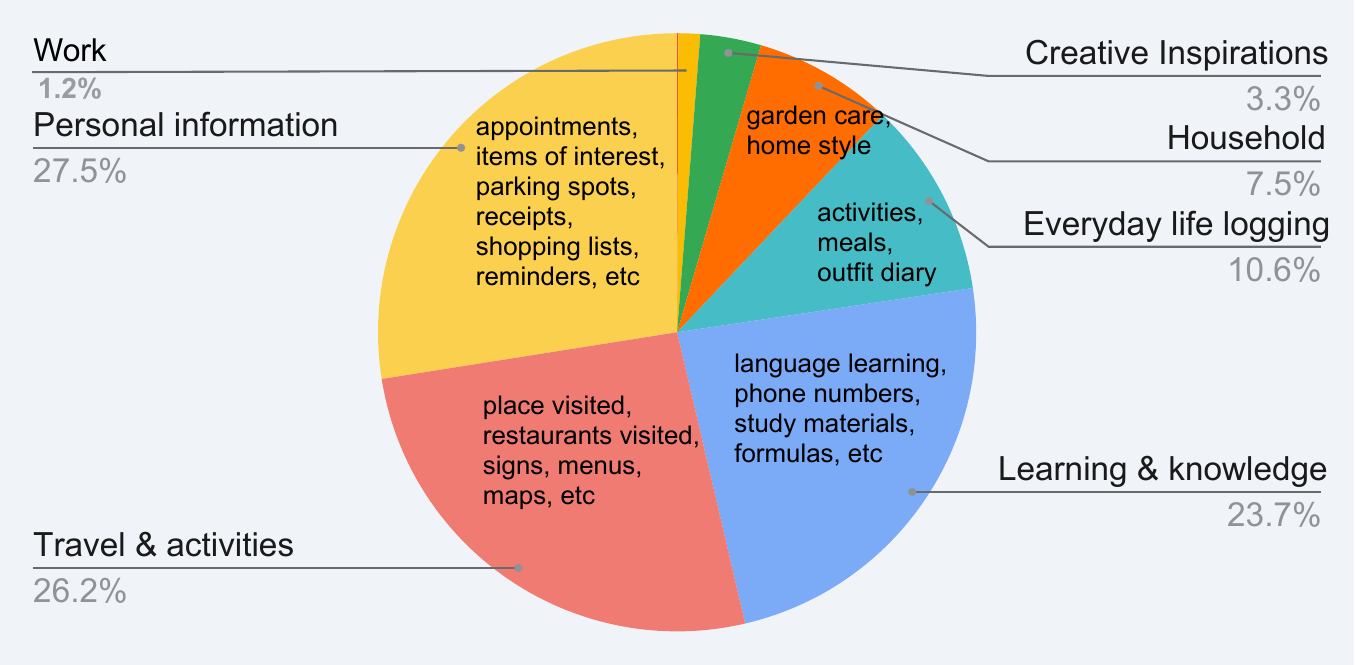}
\caption{Category distribution of the MemoryQA-test-l dataset, with detailed use cases shown within the pie chart.}
\label{fig:category}
\end{figure}

\begin{table}[t!]
    \centering
\resizebox{\columnwidth}{!}{%
    \begin{tabular}{lcccc}
       \toprule
       Dataset & \multicolumn{4}{c}{Number of questions} \\
       \midrule
       & \multicolumn{4}{c}{Temporal/location constrains} \\
       \cmidrule(lr){2-5}
       & time-only & loc.-only & time \& loc. & no-constrain \\
       $s$ & 633 & 53  & 28  & 612  \\
       $l$ & 895 & 335 & 355 & 1386 \\
       \midrule
       & \multicolumn{4}{c}{Aggregation level} \\
       \cmidrule(lr){2-5}
       & \multicolumn{2}{c}{single-memory} & \multicolumn{2}{c}{multi-memory}  \\
       $s$ & \multicolumn{2}{c}{1,326}  & \multicolumn{2}{c}{16}  \\
       $l$ & \multicolumn{2}{c}{2,797}  & \multicolumn{2}{c}{174}\\
       \bottomrule
    \end{tabular}
    }
    \caption{Statistics of the test dataset. The table reports question counts by aggregation level (single- vs. multi-memory) and by temporal/location constraints. Aggregation level indicates how many memories are relevant to a question. Temporal constraints specify a date range of relevant memories (e.g., {\em “what did I save yesterday”}), while location constraints specify where the memory was created (e.g., {\em “what was the hotel name I saved in Las Vegas”}).}
    \label{tab:distribution}
\end{table}

\subsection{Prompts}
\subsubsection{Memory Augmentation}
\textit{Invocation Completion} is generated using Llama3.2-90B-vision. Figure \ref{fig:comp-prompt} shows the complete prompt. We use it to rewrite the user invocation into a complete sentence with an explicit target.
Besides, we generate \textit{QA-guided image description} using Llama3.2-90B-vision with the prompt shown in Figure \ref{fig:qa-guided-img-desc}.

\subsubsection{Datetime matching}
\label{datetime-match-prompt}
We used a temporal matching module leveraging LLM to extract date time information from the raw user question. The prompt is shown in Figure \ref{fig:datetime-match}. 

\subsubsection{Answer Generation}
For text models, we used the prompt presented in Figure \ref{fig:ans-gen} to generate the final answer. Memories are converted to a JSON object with the following fields: memory id, invocation command, visual content (including invocation completion and image description), OCR text, creation date, and address.

For vision models, we make small changes to the prompt above. First, we added image as attachments. To handle multiple retrieved images, we concatenate all images into a single image with a 5 px spacing between images. Second, we added the following line to the instruction part of the prompt:
\begin{lstlisting}[basicstyle=\scriptsize\ttfamily, breaklines=true]
-- Input structure: If there are multiple user memories, images from all memories are concatenated together. The order of image is consistent with the order of user memories.
\end{lstlisting}

To generate the final answer along with the reference memory id list, we modify the task into 
\begin{lstlisting}[basicstyle=\scriptsize\ttfamily, breaklines=true]
Your current task is to answer questions about user memory. You need to provide two fields in JSON format, {id_list: [""], response: ""}.
\end{lstlisting}
and added the explanation of id list into the detailed instruction:
\begin{lstlisting}[basicstyle=\scriptsize\ttfamily, breaklines=true]
  -- id_list: A list of memory_id of the memories used for answering the query.
\end{lstlisting}

\subsection{Dataset}
\subsubsection{VQA-Mem Dataset}
\paragraph{Image and QA Pair Selection}
We use Llama3.3-70B to remove QA pairs that are not recall questions and answers. The prompt for data filtering is shown in Figure \ref{fig:recall-q} and Figure \ref{fig:recall-a}.
We only keep images with at least one recall QA pair, ensuring relevance for Memory-QA.

\paragraph{Invocation Commands Generation}
We use few-shot learning to prompt Llama3.2-90B to generate two different invocation commands for each selected image from the previous step. The prompt used to generate invocation commands are shown in Figure \ref{fig:invo}.

\subsubsection{MemoryQA Dataset}
\label{sec:mem-QA-dataset}
\paragraph{Test set} The test set consists of 2,789 images captured using wearable devices. We first prompt VLM to generate synthetic temporal and location metadata for each image. Then, for each image,  we craft a set of invocation commands (e.g., {\em "remember this restaurant"}) and timestamped memory recall questions (e.g., {\em "where was the restaurant I saved last week"}). The paired image, invocation command, and temporal/location metadata forms a raw memory. To create the memory context for each recall question, we combine the source memory with 10-50 additional memory samples randomly drawn from the memory pool, ensuring all memories are created before the recall timestamp. e then prompt VLM to identify relevant positive memories from this context and generate a final answer based on the selected positive candidates. In the end, human annotator review the selected positive memories and generated answers,  making adjustments as needed to ensure high-quality question-answer pairs.

\begin{figure*}
    \includegraphics[width=\linewidth]{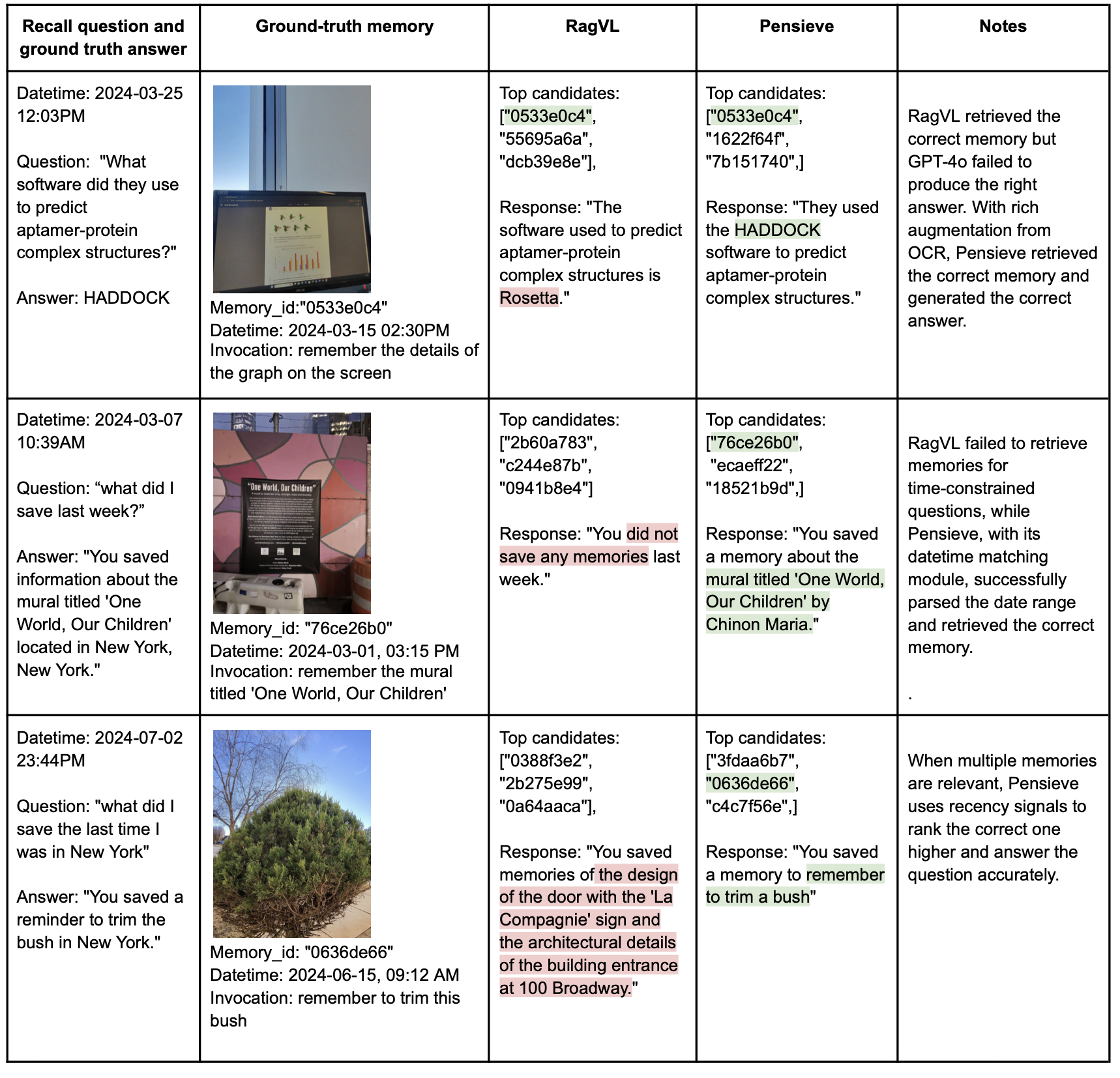}
    \caption{
        Case studies comparing the state-of-the-art RagVL with our proposed \system pipeline.
    }
    \label{fig:case_study}
\end{figure*}

Our test dataset spans diverse categories and use cases. Figure~\ref{fig:category} illustrates the distribution of the seven main categories and their associated use cases in the MemoryQA-test-l dataset. Table~\ref{tab:distribution} further summarizes the test set statistics, breaking down questions by temporal and location constraints as well as by aggregation level (i.e., whether a question involves single or multiple memories). Table~\ref{fig:case_study} provides examples that compare the performance of a state-of-the-art baseline (RagVL) with our proposed \system, highlighting differences in retrieval accuracy and answer generation.

\paragraph{Train set} Our training set follows a similar process, with two key differences. First, the invocation commands and recall questions are generated by an VLM. Second, instead of human validation, we use VLM to filter irrelevant memories from the positive candidates based on the question and generated answer.

\subsection{SOTA MM-RAG Implementation}
\subsubsection{VISTA} 
Following the methodology in \citep{zhou2024vista},  we use the vis-BGE-m3 model as our multi-modal encoder. For each memory entry $M_i$, the encoder processes the raw image, invocation command, creation datetime, and address to generate an embedding vector. At runtime, we encode the user query using vis-BGE-m3 into a query embedding. Then we compute the dot product  between the query and memory embeddings to obtain a similarity score. The top-$K$ memory entries with the highest similarity scores are selected and passed to GPT-4o for answer generation.

\subsubsection{RagVL} Our implementation of the RagVL pipeline follows the approach described in \citep{chen2024mllm}. It comprises two stages: a retriever and a vision-language model (VLM)-based re-ranker. 

In the retrieval stage, we use CLIP to encode memory images and retrieve the top 20 most relevant candidates for a given user query. These candidates are then passed to the re-ranking stage using LLaVA-v1.5-13B, which was fine-tuned on the WebQA dataset. Each memory is represented as a JSON-formatted textual passage containing the following fields: invocation command, creation datetime, and address. The re-ranker takes this passage along with the corresponding memory image and the query as input, using the following prompt (adopted from the original paper):
\begin{lstlisting}[basicstyle=\scriptsize\ttfamily, breaklines=true]
Based on the image and its caption, is the image relevant to the question? Answer 'Yes' or 'No'.
\end{lstlisting}
The probability of generating the token   `Yes' is used as the re-ranking score. The final top-$K$ ranked memories are then passed to GPT-4o for answer generation.

\subsection{Signal fusion re-ranker} 
\label{sec:reranker}
Our signal fusion re-ranker combines multiple relevance scores into a single ranking. We evaluate three fusion strategies:
\begin{itemize}
    \item Max: We rank memories by its highest score across all retrievers. Ties are broken by comparing the next highest scores.
    \item Sum: We compute and rank the sum of scores from all signals for each candidate.
    \item Learned-weight: We obtain the weight of each signal by training a linear Support Vector Classifier (RankSVM) \citep{joachims2002learning} on a small randomly sampled subset of the training data with squared hinge loss. 
\end{itemize}
As shown in Table \ref{tab:retriever_reranker}, the learned-weight approach achieved the best retrieval performance.

\subsection{Overall $A_{key}$ Results}
\begin{table}[t!]
    \centering
    \small
    \resizebox{\columnwidth}{!}{%
    \begin{tabular}{lcccc}
       \toprule
       Model & WebQA & VQA-Mem & \multicolumn{2}{c}{MemoryQA} \\
        &  &  & $s$ & $l$ \\
       \midrule
       \multicolumn{5}{c}{\textit{Baseline (w/o aug., CLIP)}} \\
       GPT-4o (vis) & 67.6 & 52.8  & 53.3 & 49.1 \\
       Llama3.2-90B (vis) & 61.0  & 48.2  & 52.5 & 50.5 \\
       Llama3.3-70B (txt) & 49.4 & 7.3  & 34.8 & 40.0 \\
       \midrule
       \multicolumn{5}{c}{\textit{SOTA MM-RAG systems, w/ GPT-4o}} \\
       VISTA &  71.0 & 52.9 & 58.2 & 57.2 \\
       RagVL & \textbf{72.2}  & 52.5 &  57.5 & 58.3 \\
       \midrule
       \multicolumn{5}{c}{\textit{\system: w/ aug., Multi-signal Retriever}} \\
       GPT-4o (vis) & 68.7  & \textbf{55.8} &  69.4 & 66.0 \\
       Llama3.2-90B (vis) & 64.1  & 52.0  & 70.0 & 65.3 \\
       Llama3.3-70B (txt) & 59.5  & 47.5 & \textbf{71.0}  & \textbf{66.1} \\
       \bottomrule
    \end{tabular}
    }
    \caption{E2E QA results $A_{key}$. \system outperforms the baseline and state-of-the-art solutions on recall questions from VQA-Mem and MemoryQA by a big margin.}
    \label{tab:overall-key}
\end{table}

E2E QA results $A_{key}$ is shown in Table \ref{tab:overall-key}. The score trends in general align with $A_{llm}$ in Table \ref{tab:overall}, demonstrating the effectiveness of \system evaluated under different quality metrics.

\subsection{Ambiguous Temporality}
To demonstrate the robustness of our system under ambiguous temporal reference, we conduct a focused case study using 10 user queries with vague temporal cues (e.g., {\em “What was the event we went to a while back?”}, {\em “What did I save after the meeting?”}, {\em “Which movie did I watch a while ago?”}). Our findings indicate that in these cases, the datetime matching module often produces an empty search date range and does not trigger the recency signal. As a result, retrieval relies primarily on semantic similarity between the query and the memory content via embeddings.

\subsection{Model fine tuning}
We fine-tuned the Llama3.1-8B-Instruct model using supervised fine-tuning (SFT) for the answer generation stage. The training was conducted for 4 epochs using the following hyperparameters: learning rate $2\times10^{-6}$, batch size 1, gradient accumulation steps 4, minimum learning rate ratio 0.1, warm-up steps 200. The best performance was achieved after the first epoch.

\begin{figure*}[ht]
\centering
\begin{PromptBox}
You are an evaluator, and you are given a task to evaluate a model predictions with a given question. Let's follow the instructions step by step to make a judgement.
1. As the first step, you need to check whether the prediction was really answering the question.
2. If the model prediction does provide a meaningful answer, judge whether the model Prediction matches the ground truth answer by reasoning according to the following steps:
2.1: Always assume the ground truth is correct. 
2.2: Pay attention to theses special cases:
    a. If the ground truth answer contains numbers,  the value of "accuracy" is true only if numbers in ground truth and numbers in model predictions match very well; in case of math questions, "accuracy" is true only if the numbers in model predictions EXACTLY matches the numbers in ground truth;
    b. If the ground truth answer contains time, and/or time range, "accuracy" is "true" only if if times and time ranges in ground truth and model predictions match very well.
    c. If the ground truth answer contains a set of objects, "accuracy" is "true" if the model prediction covers most of the objects in the ground truth; however, "accuracy" if "false" if the model prediction has a lot of objects that are not in the ground truth.
    d. If the ground truth is something similar to "I don't know", "accuracy" is "true" only if the model prediction also implies the similar thing.
2.3: Even if the prediction statement is reasonable, if it conflicts with or does not match the ground truth, "accuracy" should be "false".
2.4. "Accuracy" is true if the ground truth information is covered by the prediction. The prediction is allowed to provide more information but should not be against the ground truth. If it is hard to decide whether the prediction matches ground truth, "accuracy" should be "false".
Think step by step following the instructions above, and then make a judgment. Respond with only a single JSON blob with an "explanation" field that has your short(less than 100 word) reasoning steps and an "accuracy" field which is "true" or "false". 
Question: {{question}}
Ground truth: {{answer}}
Prediction: {{prediction}}
\end{PromptBox}
\caption{Prompt for Llama3.3-70B to auto-judge the response, where \{\{prediction\}\} is the response from answer generator.}
\label{fig:autojudge}
\end{figure*}

\begin{figure*}[ht]
\centering
\begin{PromptBox}
You are a helpful assistant that can generate evaluation data for a memory save and retrieval stack. The stack helps users to remember what they saw in the image and allows them to retrieve the memory and ask questions about it. Given an image, your task is to generate these items: 
(1) "invocation1": a user query to create the main memory of what the user sees.
(2) "invocation2": a different user query to create another memory of what the user sees.

For example,
[image is about a LG refrigerator with a price tag] 
Response:
{"invocation1": "remember the fridge",
"invocation2": "remember the price"}

Now look at the image and generate the items. 
\end{PromptBox}
\caption{Invocation commands generation prompt.}
\label{fig:invo}
\end{figure*}

\begin{figure*}[ht]
\centering
\begin{PromptBox}
## Task Description
You are a skilled assistant capable of completing invocation sentences based on an image.

### Key Definitions
* **Invocation Sentence**: A concise transcription associated with an image object, capturing key information for later recall.
* **Invocation Completion**: A completed invocation sentence with additional details about the object, such as attributes, actions, or context.

Example:
Invocation Sentence: "remember the restaurant"
Invocation Completion: "remember the Korean restaurant named 'Kochi' in NYC"

## Input
You have been provided with an invocation sentence for the image:
{{invocation}}

## Output Requirements
Please analyze the image and generate an invocation completion for the invocation sentence.
\end{PromptBox}
\caption{Prompt for VLM to rewrite the invocation into a complete sentence with explicit target. }
\label{fig:comp-prompt}
\end{figure*}

\begin{figure*}[ht]
\centering
\begin{PromptBox}
## Task Description
You are a skilled assistant capable of generating recall questions and answers based on an image, as well as creating a detailed image description that addresses all the recall questions.

### Key Definitions
* **Recall Question**: A user query about an image from their past, aiming to retrieve relevant information among all images at a later time.
  Examples:
  - What is the name of the Korean restaurant?
  - Where did I park my car today?
  - Where is the bedroom key?
  - When is the milk expiration date?
  - What vegetables do I have in my fridge?
  Non-examples:
  - What is the girl doing in the image?
  - Why are there stickers on the oranges?
  - What time is it?
  - Where is the bench located in the image?
  - What object is on the right side of the image?
* **Recall Answer**: A precise response to a recall question, enabling information recall without visual reference.

## Output Requirements
Given an image, provide the following items in JSON format:
  - 'recall_question': A list of potential recall questions a user might ask about the image.
  - 'recall_answer': A list of corresponding recall answers for each recall question.
  - 'image_description': A comprehensive image description with additional details that address all the recall questions above.

Please analyze the provided image and generate the required items.
\end{PromptBox}
\caption{QA guided image description prompt}
\label{fig:qa-guided-img-desc}
\end{figure*}

\begin{figure*}[ht]
\centering
\begin{PromptBox}
Given a user question recalling a saved memory and its timestamp, extract the search_start_date and search_end_date of the user question. Also, predict whether the user wants to search for the most recent memory or not.
- search_start_date: The start date of the search range in the database, formatted as "YYYY-MM-DD" (e.g., "2024-08-25").
- search_end_date: The end date of the search range, also formatted as "YYYY-MM-DD" (e.g., "2024-08-25").
- search_recent: A boolean value indicating whether the user wants to search for the most recent memory.
- If no time information is provided in the question, set search_start_date and search_end_date to empty strings ("").

For example:
question: "where did I park yesterday"
recall_time: "2024-05-06 Tuesday"
output:
{"search_start_date": "2024-05-05", "search_end_date": "2024-05-05", "search_recent": false}

question: "which book did I saved last time"
recall_time: "2024-08-26 Monday"
output:
{"search_start_date": "", "search_end_date": "", "search_recent": true}

Here is the user question and recall time:
question: {{question}}
recall_time: {{recall_time}}
Now generate the output in JSON format without any other text.
\end{PromptBox}
\caption{Prompt for Llama3.3-70B to parse the search date time and recency signal from the raw user query.}
\label{fig:datetime-match}
\end{figure*}

\begin{figure*}[ht]
\centering
\begin{PromptBox}
### Instruction:
You are an assistant. Your current task is to answer questions about user memory.
Here are detailed instructions:
  -- Input structure: When given a user memory, it will contain: memory_id, created_datetime, description, visual_content, ocr_text.
  -- Input structure: visual_content is a description of the image attached to the user memory, and ocr_text is the text extracted from the image. Both are optional and might not be available.
  -- Response format: Be terse and to the point, don't mention your reasoning, and answer in a single sentence.
Now look at all the content in all given user memories, and provide "response".
### Input:
Current date time is: {current_date_time}
Candidate memories: {memory_candidates}
Current turn:
- User: {user_query}
### Response:
\end{PromptBox}
\caption{Answer generator prompt where \{\{memory candidates\}\} are  memories in a JSON format.}
\label{fig:ans-gen}
\end{figure*}

\begin{figure*}[ht]
\centering
\begin{PromptBox}
You are a helpful assistant that can identify real recall questions for a memory save and retrieval stack.

Group Definition: Recall Question
Description: A query posed by a user regarding an image from their past, with the intention of retrieving associated useful information at a later time.
The following sentences belong to the group "Recall Question":
- What is the brand of the milk powder?
- What is the name of the korean restaurant?
- Where did I park my car today?
- Where is the bathroom key?
- When the monthly fees are due?
- What kind of plants do I have in my garden?
- What is my license plate number?
The following sentences do not belong to the group "Recall Question":
- What is the girl doing?
- What color is the stop light?
- Why are there stickers on the oranges?
- What time is it?
- Where is the man standing?
- What is the word that starts with 'W'?
- What object is in focus?

Does the following sentence belong to the group "Recall Question"? Answer only with "True" or "False".
\end{PromptBox}
\caption{Recall question filtering prompt.}
\label{fig:recall-q}
\end{figure*}

\begin{figure*}[ht]
\centering
\begin{PromptBox}
You are a helpful assistant that can identify good recall answers for a memory save and retrieval stack.

Group Definition: Recall Answer
Description: A precise response to a user's query about an image from their past, enabling recall of associated information without visual reference.
The following sentences belong to the group "Recall Answer":
- Question: Where are the blue container bins? Answer: under the dinning table
- Question: What phone number is on the sign? Answer: 604-909-7275
- Question: What type of condiment is on the top shelf? Answer: mayonnaise
- Question: Where is the bus parked by? Answer: main st
- Question: What brand is the bike? Answer: yamaha
The following sentences do not belong to the group "Recall Answer":
- Question: Where are the blue container bins? Answer: left
- Question: What phone number is on the sign? Answer: 0
- Question: What type of condiment is on the top shelf? Answer: condiment
- Question: Where is the bus parked by? Answer: m
- Question: What brand is the bike? Answer: no brand

Does the following sentence belong to the group "Recall Answer"? Answer only with "True" or "False".
\end{PromptBox}
\caption{Recall answer filtering prompt.}
\label{fig:recall-a}
\end{figure*}

\end{document}